\documentclass{article}

\usepackage{microtype}
\usepackage{graphicx}
\usepackage{booktabs} % for professional tables
\usepackage{url}
\usepackage{graphicx}
\usepackage{subcaption}
\usepackage{enumitem}
\usepackage{multirow}
\usepackage{booktabs}
\usepackage{wrapfig}
\usepackage[linesnumbered,lined,commentsnumbered,ruled,vlined,algo2e]{algorithm2e}
\usepackage{color, colortbl}
\usepackage{tablefootnote}
\usepackage{threeparttable}
\usepackage[most]{tcolorbox}
\usepackage{spverbatim}
\usepackage{fancyvrb}
\usepackage{fvextra}
\usepackage{xcolor}
\newtcolorbox[auto counter]{prompt}[1][]{
  enhanced,
  breakable,
  fonttitle=\scshape,
  #1
}

\usepackage{hyperref}

\usepackage[accepted]{icml2025}

\usepackage{amsmath}
\usepackage{amssymb}
\usepackage{mathtools}
\usepackage{amsthm}

\usepackage[capitalize,noabbrev]{cleveref}

\theoremstyle{plain}

\theoremstyle{definition}

\theoremstyle{remark}

\usepackage[textsize=tiny]{todonotes}

\icmltitlerunning{ELEMENTAL}

\begin{document}

\twocolumn[
\icmltitle{ELEMENTAL: Interactive Learning from Demonstrations and Vision-Language Models for Reward Design in Robotics}

% It is OKAY to include author information, even for blind
% submissions: the style file will automatically remove it for you
% unless you've provided the [accepted] option to the icml2025
% package.

% List of affiliations: The first argument should be a (short)
% identifier you will use later to specify author affiliations
% Academic affiliations should list Department, University, City, Region, Country
% Industry affiliations should list Company, City, Region, Country

% You can specify symbols, otherwise they are numbered in order.
% Ideally, you should not use this facility. Affiliations will be numbered
% in order of appearance and this is the preferred way.
\icmlsetsymbol{equal}{*}

\begin{icmlauthorlist}
\icmlauthor{Letian Chen}{yyy}
\icmlauthor{Nina Moorman}{yyy}
\icmlauthor{Matthew Gombolay}{yyy}
% \icmlauthor{Firstname4 Lastname4}{sch}
% \icmlauthor{Firstname5 Lastname5}{yyy}
% \icmlauthor{Firstname6 Lastname6}{sch,yyy,comp}
% \icmlauthor{Firstname7 Lastname7}{comp}
%\icmlauthor{}{sch}
% \icmlauthor{Firstname8 Lastname8}{sch}
% \icmlauthor{Firstname8 Lastname8}{yyy,comp}
%\icmlauthor{}{sch}
%\icmlauthor{}{sch}
\end{icmlauthorlist}

\icmlaffiliation{yyy}{School of Interactive Computing, Georgia Institute of Technology, Atlanta, United States}
% \icmlaffiliation{comp}{Company Name, Location, Country}
% \icmlaffiliation{sch}{School of ZZZ, Institute of WWW, Location, Country}

\icmlcorrespondingauthor{Letian Chen}{letian.chen@gatech.edu}
% \icmlcorrespondingauthor{Firstname2 Lastname2}{first2.last2@www.uk}

% You may provide any keywords that you
% find helpful for describing your paper; these are used to populate
% the "keywords" metadata in the PDF but will not be shown in the document
\icmlkeywords{Machine Learning, ICML}

\vskip 0.3in
]

% this must go after the closing bracket ] following \twocolumn[ ...

% This command actually creates the footnote in the first column
% listing the affiliations and the copyright notice.
% The command takes one argument, which is text to display at the start of the footnote.
% The \icmlEqualContribution command is standard text for equal contribution.
% Remove it (just {}) if you do not need this facility.

\printAffiliationsAndNotice{}  % leave blank if no need to mention equal contribution
% \printAffiliationsAndNotice{\icmlEqualContribution} % otherwise use the standard text.

\begin{abstract}
Reinforcement learning (RL) has demonstrated compelling performance in robotic tasks, but its success often hinges on the design of complex, ad hoc reward functions. Researchers have explored how Large Language Models (LLMs) could enable non-expert users to specify reward functions more easily. However, LLMs struggle to balance the importance of different features, generalize poorly to out-of-distribution robotic tasks, and cannot represent the problem properly with only text-based descriptions. To address these challenges, we propose ELEMENTAL (intEractive LEarning froM dEmoNstraTion And Language), a novel framework that combines natural language guidance with visual user demonstrations to align robot behavior with user intentions better. By incorporating visual inputs, ELEMENTAL overcomes the limitations of text-only task specifications, while leveraging inverse reinforcement learning (IRL) to balance feature weights and match the demonstrated behaviors optimally. ELEMENTAL also introduces an iterative feedback-loop through self-reflection to improve feature, reward, and policy learning. Our experiment results demonstrate that ELEMENTAL outperforms prior work by 42.3\% on task success, and achieves 41.3\% better generalization in out-of-distribution tasks, highlighting its robustness in LfD.
\end{abstract}

\section{Introduction}

Reinforcement Learning (RL) has been shown to be a powerful tool for enabling robots to perform complex tasks across a wide range of domains, from manipulation~\citep{kober2013reinforcement, levine2016end} to navigation~\citep{tai2017virtual, zhu2017target}. However, the effectiveness of RL hinges on the availability of a carefully designed reward function that accurately encapsulates the desired behavior. Without sophisticated reward functions, RL agents often struggle to learn competent policies~\citep{10.1007/11840817_87}; worse yet, poorly designed reward functions can lead RL agents to undesirable outcomes~\citep{gupta2024behavior}. \citet{booth2023perils} shows reward specification is non-trivial even for experts, and designing reward functions that align with end-users' expectations is particularly challenging due to the varied and latent user preferences~\citep{abouelazm2024review}.

Considering recent advancements in large models (e.g., large language models (LLMs) and vision-language models (VLMs)) on text understanding~\citep{bommasani2021opportunities,touvron2023llama} and emergent abilities~\citep{kojima2022large, wei2022emergent}, researchers have explored utilizing LLMs for reward engineering. For instance, the EUREKA framework takes as input a text description of the task, queries LMs for a draft of the reward function, and trains a policy with the reward function~\citep{ma2023eureka}. This paradigm, while promising, presents several limitations. First, describing complex robotic tasks purely through language is imprecise: humans often have latent, unspoken preferences that are difficult to articulate fully, leading to incomplete or ambiguous descriptions of their objectives~\citep{nisbett1977telling, ericsson1980verbal, hoffman1995eliciting, feldon2007implications}. Second, even if all objective function components are accurately conveyed, determining the relative importance or weights of these components poses another significant challenge: assigning these weights involves subtle mathematical trade-offs, something that LLMs are not particularly equipped to handle. As a result of these limitations, methods like EUREKA struggle to generalize well to out-of-distribution tasks.

Given these limitations, a more natural and effective approach is for users to provide demonstrations of the desired behavior to supplement a general task description. Demonstrations offer rich, illustrative information that can capture not only the task objectives but also the nuanced, latent preferences that may be difficult to express verbally. Learning from Demonstration (LfD) approaches seek to leverage human-provided demonstrations to reverse-engineer the underlying objective and optimize a policy accordingly~\citep{chen2020joint, corl2022_flair, suay2016learning, ravichandar2020recent}. However, a key challenge in LfD is the ambiguity in interpreting demonstrations -- there can be an infinite number of possible reward functions that could explain the same set of demonstrations, a problem commonly referred to as the \emph{reward ambiguity problem} in Inverse Reinforcement Learning (IRL)~\citep{abbeel2004apprenticeship}. Prior methods have sought to address this ambiguity by pre-designing features to constrain the space of possible reward functions, but this often limits the flexibility and generalizability of the learned rewards and policies~\citep{robotics7020017,arora2021survey}. Our key insight is that language models are well-suited to contextualize demonstrations and infer task features, narrowing down the possible interpretations and enabling robots to learn more robustly. 

We propose to integrate the strengths of language models and LfD methods, leveraging (1) the emergent reasoning capabilities of language models to identify robust and relevant objective function components, and (2) the demonstration-matching capabilities of LfD to determine the optimal weighing of these components. Crucially, we incorporate visual demonstrations into Vision-Language Models (VLMs), facilitating a more comprehensive understanding of human objectives. Additionally, we introduce a self-reflection mechanism that enables VLM and LfD to iteratively improve both the feature extraction and the reward \& policy learning. This fusion of LfD and VLMs offers a novel pathway for more effective robotic learning from demonstrations. 
This paradigm also mirrors how humans naturally learn from others. When observing a demonstration, humans typically (1) identify the key aspects of the task, (2) formulate a policy to match the demonstration based on those salient features, and (3) reflect on the discrepancies between their own behavior and the demonstration to refine their understanding and execution~\citep{locke1987social,di2014learning}. By iterating through these steps, humans progressively improve their performance. This iterative cycle of observation, reflection, and refinement is not only fundamental to human learning but also serves as an ideal framework for robotic learning~\citep{chernova2014robot}.

To develop this novel integration between VLM and LfD, we introduce ELEMENTAL (intEractive LEarning froM dEmoNstraTion And Language). ELEMENTAL enables robots to identify key task features from human demonstrations, learn rewards and policies that align on these features, and iteratively reflect on their performance to improve over time. Our key contributions are three-folds:\vspace{-10pt}
\begin{enumerate}[leftmargin=*]
    \item We propose a novel, general framework that integrates VLMs and LfD and introduces an iterative self-reflecting mechanism for autonomous performance improvement. ELEMENTAL is the first to incorporate visual demonstration inputs into language models to accomplish LfD, which facilitates more accurate behavior understanding.\vspace{-7pt}
    \item We evaluate ELEMENTAL on a set of challenging, standard robotic benchmarks in IsaacGym, demonstrating its superior performance over previous state-of-the-art (SOTA) reward design and LfD methods by 42.3\%, showcasing its effectiveness.\vspace{-7pt}
    \item We further assess ELEMENTAL’s generalization capabilities by designing novel variants of the standard benchmarks. Our results show that ELEMENTAL achieves 41.3\% better generalization than existing methods, underscoring the importance of combining VLMs with LfD.\vspace{-7pt}
\end{enumerate}
\begin{figure*}[t]
    \centering
    \includegraphics[width=0.9\linewidth]{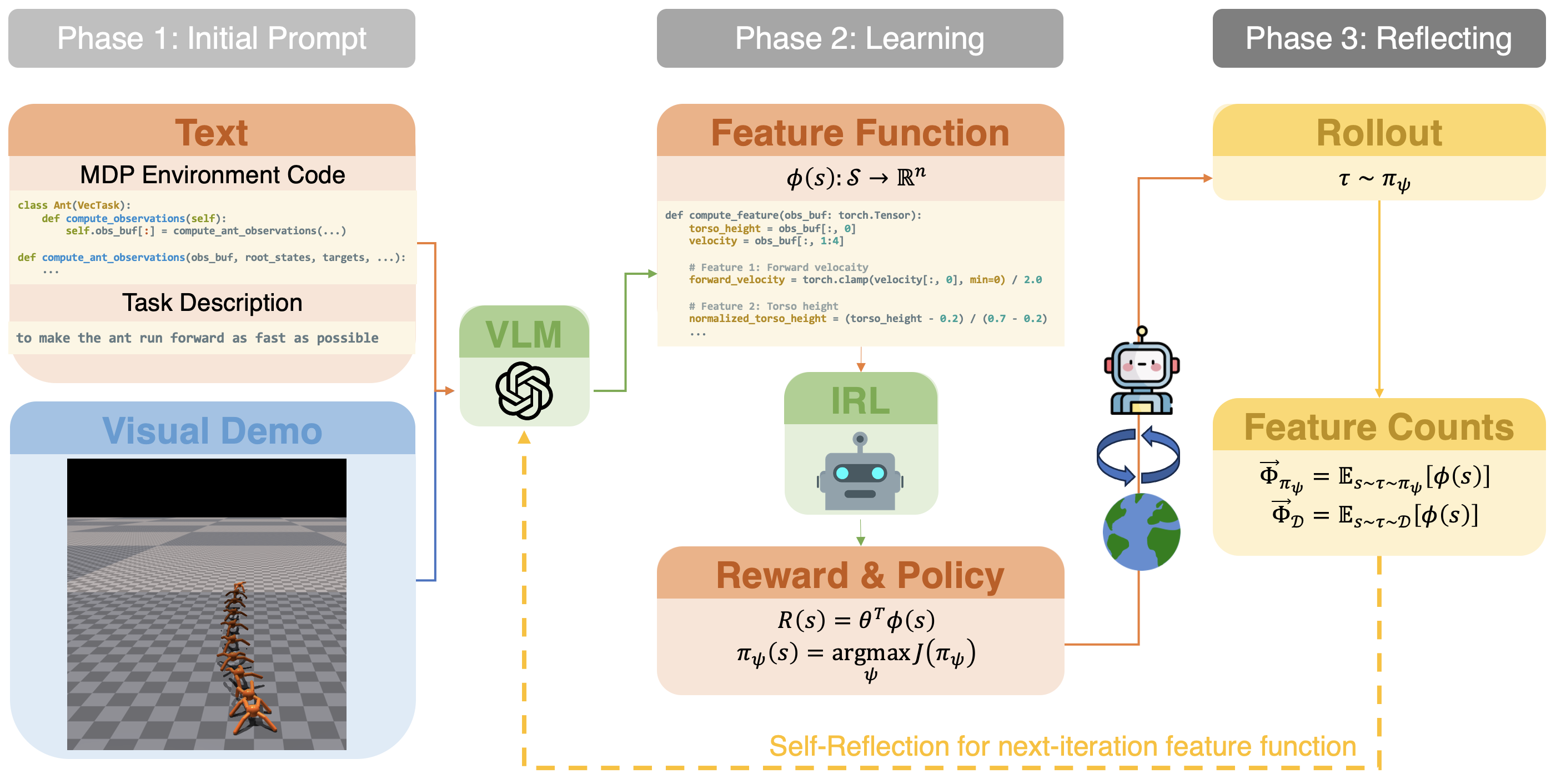}
    \caption{This figure illustrates the overall pipeline of ELEMENTAL. The process begins with an initial prompt to the VLM, which generates a draft of the feature function based on both textual descriptions and visual demonstrations. In the learning phase, ELEMENTAL infers the reward and policy from the drafted feature function and the demonstration. In the final phase, ELEMENTAL performs self-reflection by comparing the feature counts from the generated trajectory and the demonstration, again utilizing the drafted feature function. This self-reflection loop updates the feature function by feeding the results back to the VLM for iterative refinement.}\vspace{-10pt}
    \label{fig:ELEMENTAL-overview}
\end{figure*}
\vspace{-10pt}
\section{Related Work}
\paragraph{Learning from Demonstration (LfD) --} LfD approaches, such as Behavior Cloning (BC) and IRL, have long been used to enable robots to learn from human-provided demonstrations. BC~\citep{ross2011reduction}, a supervised learning approach, is effective for relatively simple tasks, but it is prone to compounding errors (known as \textit{covariate shift}). IRL~\citep{ng2000algorithms, abbeel2004apprenticeship, ziebart2008maximum, ziebart2010modeling} seeks to infer the underlying reward function that explains the demonstrated behavior. However, reward ambiguity poses a significant challenge -- an infinite number of reward functions could explain the same behavior. This challenge becomes exacerbated in complex domains with limited, hetergeneous demonstrations~\cite{chen2020joint, peng2024preference}. ELEMENTAL addresses this key limitation by integrating VLMs to inject emergent reasoning capabilities into the learning process. VLMs reduce ambiguity by providing semantic context that allows robots to better understand relevant task features. 
% This multimodal approach enables ELEMENTAL to infer more accurate task specifications, even in environments where human demonstrations are limited or task descriptions are unclear.
\vspace{-10pt}
\paragraph{Language Models as Reward Engineers --} 
Recent works, such as EUREKA~\citep{ma2023eureka} and L2R~\citep{yu2023language}, leverage LLMs to convert language descriptions into reward functions, offering a promising alternative to manual reward engineering. However, these methods are limited by their reliance solely on brief task descriptions, restricting their ability to capture the full complexity of robotic tasks and the subtle preferences of users. Additionally, determining the appropriate weighting of different objective function components is particularly challenging, as the reward design process is disconnected from policy training, resulting in poor out-of-distribution generalization. ELEMENTAL addresses these limitations by integrating IRL with VLMs and supplementing task descriptions with demonstrations. In ELEMENTAL, the responsibility of assigning reward component weights is shifted from the VLM to IRL, which matches the reward components to the demonstrated behaviors. This allows the VLM to focus on its strength—semantic understanding and task feature identification. 

A closely related line of research is that by ~\citet{peng2024adaptive}, which also integrates feature design using LLMs with IRL. While this prior work is promising, it operates under several limiting assumptions. First, it assumes that demonstrations can be provided as input exclusively as text-based state-vector sequences. Due to this constraint, their approach is only applied to simpler tasks (e.g., problem horizons of fewer than five steps) with all but one or two features of the ground-truth features already specified. Additionally, as seen in other works~\citep{yu2023language,kwon2023reward}, this method relies on specially designed prompts tailored to each experimental domain. These limitations hinder its scalability to more complex, high-dimensional tasks typically encountered in real-world robotic applications. In contrast, ELEMENTAL leverages visual modality for demonstration inputs, making it well-suited for more complex tasks where textual descriptions alone are insufficient. In addition, to account for the complex domains, we introduce an upgraded MaxEnt-IRL approach, detailed in Sec.~\ref{subsec:learning}. By incorporating visual modality, using a general prompt across domains, and developing enhanced IRL, ELEMENTAL provides a more scalable solution, as demonstrated in our results on standard robotic benchmarks without requiring any prior knowledge of task features.
\vspace{-10pt}
\paragraph{LM-Assisted Robot Learning --} Several recent works have sought to leverage LMs to assist in robotic learning, such as RL-VLM-F~\citep{wang2024rl}, RoboCLIP~\citep{sontakke2024roboclip}, and~\cite{du2023vision}. While promising, these methods depend on a surrogate reward derived from the LM’s understanding of the task-state alignment, which can introduce inaccuracies. Furthermore, these approaches lack interactivity with users, a key component shown to be helpful in gaining human trust~\citep{chi2024interactive}. In contrast, ELEMENTAL potentially allows engineers and users to interactively refine the robot’s behavior, ensuring that the robot is user-aligned.

Other works, such as \citet{wang2023prompt} and \citet{mahadevan2024generative}, directly query LMs to output robot actions or primitives. These approaches rely heavily on the LLM’s ability to plan and optimize actions, but LLMs are not inherently designed for mathematical optimization required for robot control. ELEMENTAL addresses these limitations by using VLMs to understand task features and deferring the demonstration-to-policy alignment to IRL algorithms, which are better suited to optimize behavior.

\begin{figure*}[t]
    \centering
    \begin{minipage}[t]{0.5\textwidth}
        \vspace{0pt}
        \begin{subfigure}[t]{\textwidth}
            \includegraphics[height=0.9\textwidth]{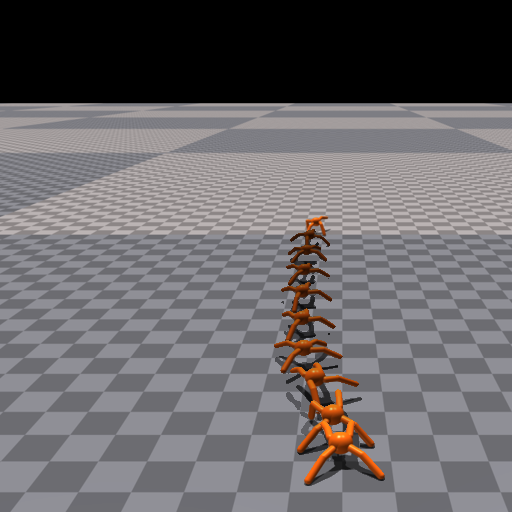}
            \caption{Superimposing ant demonstration.}
            % \label{fig:topleft}
        \end{subfigure}%
    \end{minipage}%
    \hfill
    \begin{minipage}[t]{0.48\textwidth}
        \vspace{0pt}
        \setcounter{subfigure}{1}
        \begin{subfigure}[t]{0.46\textwidth}
            \includegraphics[width=\textwidth,height = 0.9\textwidth]{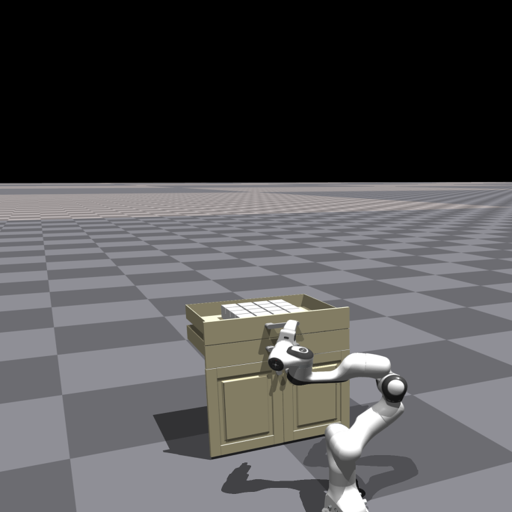}\vspace{-3pt}
            \caption{start of the trajectory. \vspace{-5pt}}
            % \label{fig:topleft}
        \end{subfigure}%
        \hfill
        \begin{subfigure}[t]{0.46\textwidth}
            \includegraphics[width=\textwidth,height = 0.9\textwidth]{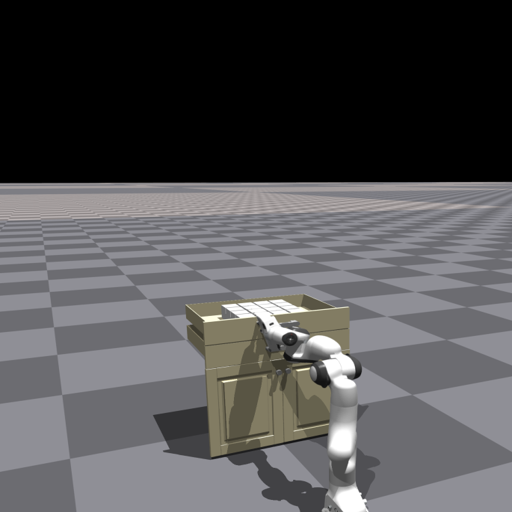}\vspace{-3pt}
            \caption{reaching the cabinet handle. \vspace{-5pt}}
            % \label{fig:topright}
        \end{subfigure}
        
        \vspace{7pt}  % This creates equal vertical spacing
        
        \begin{subfigure}[t]{0.46\textwidth}
            \includegraphics[width=\textwidth,height = 0.9\textwidth]{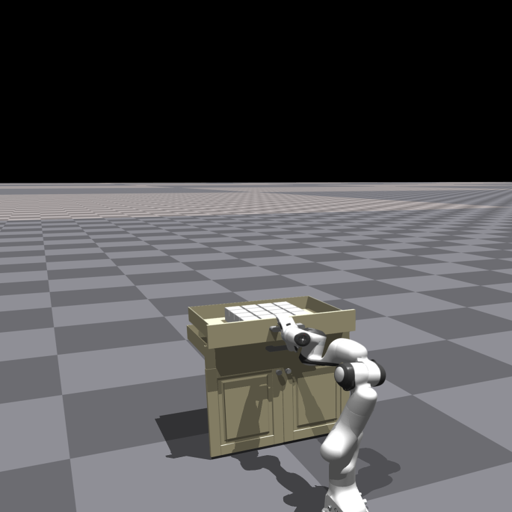}\vspace{-3pt}
            \caption{opening the cabinet. }
            % \label{fig:bottomleft}
        \end{subfigure}%
        \hfill
        \begin{subfigure}[t]{0.46\textwidth}
            \includegraphics[width=\textwidth,height = 0.9\textwidth]{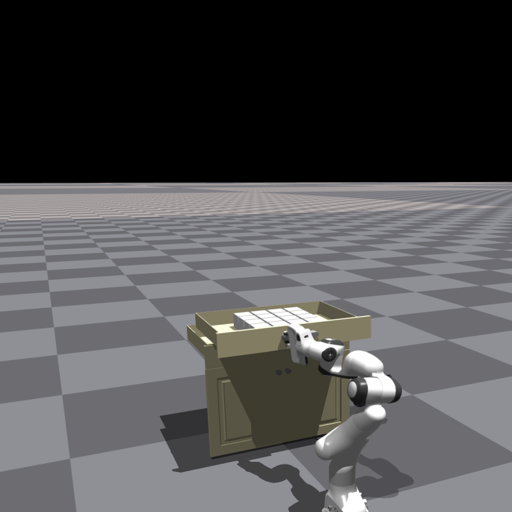}\vspace{-3pt}
            \caption{the cabinet is opened.}
            % \label{fig:bottomright}
        \end{subfigure}
    \end{minipage}
    \vspace{-10pt}
    \caption{This figure illustrates the visual demonstrations for both locomotion and manipulation tasks. (a) shows an example from the Ant locomotion task, where superimposed images are used. For manipulation tasks, superimposed images can result in cluttered robot poses, so we use key frames as visual demonstration inputs instead. (b-e) present four key frames from the FrankaCabinet manipulation task.}\vspace{-5pt}
    \label{fig:visual_demo_illustration}
\end{figure*}

\vspace{-10pt}
\section{Preliminary}
\label{sec:preliminary}

In this section, we introduce Markov Decision Process, Inverse Reinforcement Learning, and Maximum-Entropy IRL. 

\emph{\textbf{Markov Decision Process}}:
We formulate the robot learning problem as a Markov Decision Process (MDP), defined by the tuple $(\mathcal{S}, \mathcal{A}, T, R, \gamma)$. $\mathcal{S}$ is the set of states the agent can occupy. $\mathcal{A}$ is the set of actions the agent can take. $T: \mathcal{S} \times \mathcal{A} \times \mathcal{S} \rightarrow [0,1]$ represents the transition probability function, where $T(s, a, s')$ gives the probability of transitioning from state $s$ to state $s'$ after taking action $a$. $R: \mathcal{S} \rightarrow \mathbb{R}$ is the reward function that assigns a scalar reward to each state. $\gamma \in [0, 1)$ is the temporal discount factor. The goal of Reinforcement Learning (RL) is to learn a policy $\pi: \mathcal{S} \rightarrow \mathcal{A}$ that maximizes the expected cumulative reward, given by: $J(\pi) = \mathbb{E}_{\tau \sim \pi}\left[\sum^\infty_{t=0}\gamma^t R(s_t)\right]$. 

\emph{\textbf{Inverse Reinforcement Learning}}: In IRL, instead of explicitly programming a robot's behavior, we aim to learn the underlying objective or reward function, $R(s)$, which explains the behavior demonstrated by a human expert. A set of demonstrations are given, $\mathcal{D} = \{ \tau_i \}_{i=1}^N$, where each trajectory $\tau_i = (s_1^i, a_1^i, s_2^i, a_2^i, \dots)$ consists of a sequence of states and actions. We assume that the true reward function, $R(s)$, is a linear combination of features: $R(s)=\theta^T\phi(s)$, where $\phi(s) \in \mathbb{R}^d$ is a feature vector representing the task-relevant or user-preference-relevant properties of the state, and $\theta \in \mathbb{R}^d$ is a weight vector that specifies the relative importance of each feature. The goal of IRL is to recover $\theta$ based on the provided demonstrations, $\mathcal{D}$. MaxEnt-IRL~\citep{ziebart2008maximum} models the likelihood of a trajectory under the assumption that the expert's behavior is stochastically optimal, as shown in Eq.~\ref{eq:maxent_irl_prob}. In this equation, $Z(\theta)$ is the partition function, $Z(\theta)=\sum_\tau\exp\left(\sum_{s_t\in\tau}R(s_t)\right)$. 
\par\nobreak{\parskip0pt \noindent
\begin{align}
    P(\tau|\theta)&=\frac{1}{Z(\theta)}e^{\sum_{s_t\in\tau}R(s_t)} =\frac{1}{Z(\theta)}e^{\sum_{s_t\in\tau}\theta^T\phi(s)} \label{eq:maxent_irl_prob}
    \vspace{-10pt}
\end{align}
}MaxEnt IRL seeks to infer the reward weights $\theta$ that maximize the likelihood of the expert demonstrations (Eq.~\ref{eq:maxent_irl}). 
\par\nobreak{\parskip0pt \noindent
\begin{align}
\label{eq:maxent_irl}
\hat{\theta}=\arg\max_{\theta}\sum_{\tau\in\mathcal{D}}{\log P(\tau|\theta)}\vspace{-5pt}
\end{align}}
In ELEMENTAL, we upgrade MaxEnt-IRL to be suitable for high-dimensional robotic tasks, allowing us to link features from VLMs to demonstrations via weight vector $\theta$.

\section{Method}
In this section, we present ELEMENTAL, which consists of three interconnected phases (Figure~\ref{fig:ELEMENTAL-overview}). The first phase involves constructing an initial feature function through a VLM based on visual human demonstrations and environment specifications (Sec.~\ref{subsec:initial_prompt}). In the second phase, this feature function is integrated with IRL to learn a reward function and policy that best align with the demonstrated behaviors (Sec.~\ref{subsec:learning}). The final phase introduces a self-reflection mechanism that automatically compares the learned behavior with the demonstrations, enabling iterative refinement of the feature function, reward function, and policy (Sec.~\ref{subsec:reflecting}).

\vspace{-5pt}

\subsection{Phase 1: Initial Prompt}
\label{subsec:initial_prompt}
The first phase begins with an initial prompt of three inputs: (1) the MDP environment code specifying the environment's state space $\mathcal{S}$, (2) a text description of the task, and (3) a visual human demonstration of the desired task. The form of the visual demonstration depends on the task domain (Figure~\ref{fig:visual_demo_illustration}): for tasks where superimposition is meaningful (e.g., locomotion tasks such as the moving-forward ant or humanoid), we compose a superimposed image that displays the sequence of motions from demonstration (shading indicates temporal progression). For manipulation tasks, where superimposition can result in cluttered robot poses (e.g., FrankaPanda Cabinet), four keyframes picked by robotic experts are provided to the VLM to illustrate stages of the task execution. By incorporating visual demonstrations alongside language-based task descriptions, ELEMENTAL allows the VLM to generate feature functions that more accurately reflect the user’s latent goals. 

These inputs are then processed by the VLM, which leverages its emergent capabilities to infer task-relevant features. The goal is not just to describe the state but also to capture important factors that align with the user's intentions for the task. The output of this phase is a feature function $\phi: \mathcal{S} \rightarrow \mathbb{R}^n$, where $\phi(s) = (\phi_1(s), \phi_2(s), \dots, \phi_n(s))$ is an $n$-dimensional vector of features describing the key aspects of the task. Given VLM coding capabilities~\citep{piccolo2023evaluating}, we ask the VLM to output feature functions as Python code, providing a structured representation. If the returned code is not executable (e.g., wrong function signature), we re-prompt the VLM up to three times with the trackback information. A successfully executable feature function serves as the foundation for reward learning and policy optimization in the subsequent phase.
\setlength{\textfloatsep}{0pt}
\begin{algorithm2e}[t]
\caption{Approximate MaxEnt-IRL}
\label{algo:approxiamte_maxent_irl}
\SetKwInOut{Input}{Input}\SetKwInOut{Output}{Output}
\Input{feature function $\phi: \mathcal{S}\rightarrow\mathbb{R}^n$, demonstration $\mathcal{D}$, number of IRL iterations $m$, learning rate $\alpha$, policy learning steps $k$}
Initilize reward function feature weights $\theta=\{1/n\}_{i=1}^n$ and policy weights $\psi$. \\
\For{$i\gets 1$ \KwTo $m$}{
    \For{$j\gets 1$ \KwTo $k$}{
        Optimize $\pi_\psi$ based on $J(\pi_\psi)$ with $R_\theta(s)=\theta^T\phi(s)$ via PPO
    }
        Obtain $\nabla_\theta$ by Eq.~\ref{eq:approximate_maxent_irl_gradient} and normalize to get $\nabla_\theta'$ by Eq.~\ref{eq:normalize_gradient}\\
    Update $\theta$ by $\theta \leftarrow \theta + \alpha \nabla_\theta'$\\
    Normalize $\theta$ by Eq.~\ref{eq:normalize_theta}
}
\Output{$R_\theta, \pi_\psi$}
\end{algorithm2e}

\subsection{Phase 2: Learning}
\label{subsec:learning}
After the feature function has been drafted in Phase 1, we next optimize the reward function, $R_\theta(s)=\theta^T\phi(s)$, to match the demonstrations using the feature function, $\phi(s)$. As in Sec.~\ref{sec:preliminary}, we modify the MaxEnt-IRL algorithm to accomplish reward and policy learning in more complex tasks. We name it \emph{Approximate MaxEnt-IRL} (Algorithm~\ref{algo:approxiamte_maxent_irl}).

Directly computing the partition function $Z(\theta)$ is intractable due to the summation over all possible trajectories. We circumvent the need for explicit computation of $Z(\theta)$ with the key insight~\citep{ziebart2008maximum} that the gradient of the MaxEnt IRL objective is given by the feature expectation difference between the demonstrated trajectories and the stochastically optimal trajectory under $\theta$, as shown in Eq.~\ref{eq:maxent_irl_gradient}. 
\par\nobreak{\parskip0pt \noindent
\begin{align}
\label{eq:maxent_irl_gradient}
    \nabla_\theta=\mathbb{E}_{\tau\sim \mathcal{D}}\left[\sum_{s\in\tau}\phi(s)\right]-\mathbb{E}_{\tau\sim P(\tau|\theta)}\left[\sum_{s\in\tau}\phi(s)\right]\vspace{-5pt}
\end{align}}
We approximate $P(\tau|\theta)$ by a parameterized policy, $\pi_\psi$, that optimizes the estimated reward, $R_{\theta}(s)=\theta^T\phi(s)$,  in Eq.~\ref{eq:approximate_maxent_irl_gradient}. \par\nobreak{\parskip0pt \noindent
\begin{align}
\label{eq:approximate_maxent_irl_gradient}
    \nabla_\theta\approx\mathbb{E}_{\tau\sim \mathcal{D}}\left[\sum_{s\in\tau}\phi(s)\right]-\mathbb{E}_{\tau\sim \pi_\psi}\left[\sum_{s\in\tau}\phi(s)\right]\vspace{-5pt}
\end{align}}
In particular, we optimize the policy $\pi_\psi$ interleaved with $R_\theta$, following a paradigm similar to that of AIRL~\citep{fu2017learning}, as shown in Algorithm~\ref{algo:approxiamte_maxent_irl} lines 3-4 and line 6. Intuitively, by applying this gradient, we refine $\theta$ to ensure that the reward function $R_\theta(s)$ more accurately reflects the features emphasized in the demonstration and guide $\pi_\psi$ to align the policy's behavior closer to the demonstrated behavior. The learned policy, $\pi_\psi$, is also a natural by-product of this process, serving as the ultimate goal of LfD and laying the foundation for the reflection phase in Phase 3.

As the feature's numerical magnitude varies, we apply a normalization procedure to the reward gradient to ensure stable learning. First, we normalize the L1-norm of the gradient vector as in Eq.~\ref{eq:normalize_gradient}. 
\par\nobreak{\parskip0pt \noindent
\begin{align} 
\label{eq:normalize_gradient}
    \nabla_\theta' = \frac{\nabla_\theta}{||\nabla_\theta||_1}
    \vspace{-5pt}
\end{align}}
Next, we apply gradient ascent on the reward weight vector, $\theta$, using a learning rate $\alpha$: $\theta \leftarrow \theta + \alpha \nabla_\theta'$. After each update, we normalize $\theta$ (Eq.~\ref{eq:normalize_theta}) to stabilize the training of the policy, $\pi_\psi$, as the semantics of a reward function do not change by scaling~\citep{fu2017learning}. 
\vspace{-5pt}
\begin{align} 
\label{eq:normalize_theta}
\theta \leftarrow \frac{\theta}{||\theta||_1}
\vspace{-10pt}
\end{align}
% This normalization helps control the magnitude of the weight updates and ensures that the reward and policy optimization remains stable. By iterating the reward updates, reward normalization, and policy learning steps, ELEMENTAL optimizes both the reward function and the policy over a fixed number of iterations.

\subsection{Phase 3: Self-Reflecting on Features}
\label{subsec:reflecting}
The third phase introduces a self-reflection mechanism, designed to close the loop and iteratively refine the feature function drafted in Phase 1. After Phase 2, the learned policy, $\pi_\psi(s)$, is executed in the environment, and its behavior is compared to the demonstration, $\mathcal{D}$, based on the drafted feature function $\phi(s)$. Specifically, we calculate the expected feature counts of the generated trajectories under the current policy and for the demonstration trajectories (Eq.~\ref{eq:feature_counts}). 
\begin{align} 
    \vec{\Phi}_{\pi_\psi}=\mathbb{E}_{\tau \sim \pi_\psi} \left[ \sum_{s \in \tau} \phi(s) \right] , \vec{\Phi}_{\mathcal{D}}&=\mathbb{E}_{\tau \sim \mathcal{D}} \left[ \sum_{s \in \tau} \phi(s) \right]  \label{eq:feature_counts}
    \vspace{-10pt}
\end{align}
Discrepancies between the two feature counts indicate that the current feature function may not fully capture the relevant aspects of the task as demonstrated. 

The two feature count vectors are then fed back to the VLM, which uses the feature count differences to revise the feature function $\phi(s)$. By accounting for previously overlooked or misinterpreted features, the VLM’s understanding of the task becomes progressively more aligned with the demonstrated behavior. This process of self-reflection continues iteratively, alternating between Phase 2 (reward function and policy optimization) and Phase 3 (feature refinement), allowing the robot to improve its behavior over time.

The reflecting phase is fully automatic, leveraging the policy, $\pi_\psi(s)$, and the feature function, $\phi(s)$, generated in previous phases. As both the policy and the feature function are available, ELEMENTAL can continuously refine its understanding of the task without additional input from a human. However, should the user wish to intervene and provide further feedback or corrections, ELEMENTAL could accommodate interaction in future work through prompting.

% As the feature function evolves, ELEMENTAL generates increasingly refined reward functions and policies, ensuring continuous improvement in task performance. The result is a system that becomes progressively better aligned with user preferences, achieving a high level of task performance through iterative learning.

\vspace{-5pt}
\section{Results}
In this section, we evaluate the performance of ELEMENTAL on challenging robotic tasks from IsaacGym~\citep{makoviychuk2021isaac} against SOTA baselines (Sec.~\ref{subsec:benchmarking_result}) and show generalization to task variants (Sec.~\ref{subsec:generalization_result}).

\textbf{Environments and Tasks --}
In benchmark experiments, we test ELEMENTAL and baseline algorithms on nine challenging IsaacGym Robotics tasks, using GPU-accelerated training to enable efficient experiments. These tasks span various domains, including locomotion and manipulation, and are recognized for their complexity in the robot learning community~\citep{makoviychuk2021isaac}. For all methods that utilize a LLM/VLM, we use the OpenAI GPT-4o model unless otherwise noted. We use five demonstrations for each task collected with RL-trained policies. 

To the best of our knowledge, this is the first successful application of IRL to IsaacGym -- with or without large models -- due to its high-dimensional state and action spaces. The performance of ELEMENTAL in these tasks demonstrates its robustness and scalability, making it a suitable framework for real-world robotics problems using IRL. 

\vspace{-5pt}
\interfootnotelinepenalty=10000
\begin{table*}[t]
\centering
\begin{threeparttable}
\footnotesize
\caption{This table shows benchmarking results on IsaacGym tasks. Bold denotes the best performance except the GT Reward. }
\label{tab:benchmark}
\begin{tabular}{lccccccccc}
\toprule
\multirow{2}{*}{Method} & \multicolumn{9}{c}{IsaacGym Environments} \\
\cmidrule(lr){2-10}
   & \multirow{ 2}{*}{Cartpole}  & Ball & \multirow{ 2}{*}{Quadcopter} & Franka & \multirow{ 2}{*}{Ant} & \multirow{ 2}{*}{Humanoid} & \multirow{ 2}{*}{Anymal} & Allegro& Shadow \\
    &  & Balance &  & Cabinet &  &  &  & Hand & Hand \\
\midrule
Random (LB) & 25.42 & 87.39 & -1.63 & 0.00 & 0.00 & -0.04 & -2.45 & 0.00 & 0.02 \\
\midrule
 BC & 149.85 & 344.55 & -1.19 & 0.01 & -0.05 & -0.43 & -2.14 & 0.04 & 0.03 \\
 IRL & 28.15 & 162.06 & -1.87 & 0.00 & 0.88 & 2.13 & -2.22 & 0.01 & 0.01 \\
\midrule
EUREKA & 215.91 & 454.18 & \textbf{-0.22} & 0.21 & 6.88 & 3.78 & -4.24 & 11.12 & 0.00\tnote{1} \\
\midrule
ELEMENTAL (Ours) & 233.92 & \textbf{464.40} & -0.30 & \textbf{0.36} & \textbf{8.49} & \textbf{4.70} & \textbf{-0.83} & \textbf{22.97} & \textbf{2.71} \\
w/o Self-Reflection & 114.66 & 153.52 & -0.93 & 0.00 & 5.05 & 3.65 & -1.71 & 0.02 & 0.03 \\
w/o Gradient Normalization & 186.52 & 423.78 & -1.20 & 0.02 & 7.29 & 2.73 & -0.95 & 13.39 & 2.32 \\
w/o Weight Normalization & 192.51 & 459.33 & -0.54 & 0.02 & 3.23 & 4.87 & -1.15 & 0.04 & 1.57 \\
w/o Visual Input & 178.68 & 304.58 & -1.01 & 0.00 & 8.16 & 4.49 & -1.41 & 18.52 & 0.03 \\
w/ Text Demo\tnote{2} & 207.51 & 412.17 & -0.92 & 0.00 & 7.43 & 4.60 & -0.88 & 7.07 & 0.04 \\
w/ Random Visual Demo. & \textbf{269.46} & 352.53 & -1.07 & 0.02 & 7.13 & 3.93 & -1.07 & 20.75 & 2.33 \\
\midrule
GT Reward (UB) & 260.14 & 461.90 & -0.27 & 0.40 & 7.00 & 5.07 & -0.03 & 23.70 & 0.15 \\
\bottomrule
\end{tabular}
\begin{tablenotes}
    \item[1] We tried running with six seeds, but Eureka with GPT-4o failed to generate any executable reward function for ShadowHand. In the calculation of percentage improvements over Eureka, we treat the improvements on ShadowHand to be $100\%$.
    \item[2] As the original implementation in~\cite{peng2024adaptive} is not available, we implement the demonstration in text form with ELEMENTAL.
\end{tablenotes}
\end{threeparttable}
% \vspace{-5pt}
\end{table*}

\textbf{Baselines --}
We compare ELEMENTAL against key LfD and LLM-powered reward engineering approaches:
\vspace{-10pt}
\begin{enumerate}[leftmargin=10pt]
    \item LfD methods -- We include standard LfD techniques, i.e.~Behavior Cloning (BC) and IRL. These baselines learn from demonstration but do not incorporate the VLM feature-extraction or visual input. %This comparison demonstrates how traditional LfD methods perform on challenging tasks without access to the feature inference capabilities of VLM.
    \vspace{-7pt}
    \item EUREKA -- This is the previous SOTA method for reward design with RL in IsaacGym. EUREKA relies on LLMs to infer task features from textual inputs but does not utilize demonstrations, visual inputs, or IRL. \vspace{-7pt}
    \item Random Policy -- A random policy establishes an approximate lower bound for competent task performance.\vspace{-7pt}
    \item Ground-Truth (GT) Reward -- The performance of a policy trained with GT reward predefined in IsaacGym provides a upper bound of the task performance. Note that although we call it ``upper bound'', it is not necessarily the maximum performance one can achieve, as the RL is given the same budget of environment steps to train, and the GT reward may not be the best.\vspace{-7pt}
    \item Ablations of ELEMENTAL -- To better understand the contribution of each component in ELEMENTAL, we conduct ablation studies across six variants: 1) Without (w/o) Self-Reflection: This variant removes the self-reflection mechanism introduced in Phase 3, keeping only the VLM-guided feature inference and policy learning phases; 2) w/o Gradient Normalization: This ablation removes the gradient normalization in Eq.~\ref{eq:normalize_gradient}; 3) w/o Weight Normalization: This ablation removes the weight normalization step in Eq.~\ref{eq:normalize_theta}; 4) w/o Visual Input: This variant removes the visual demonstration input to the VLM, leaving only language-based task descriptions for feature inference; 5) w/ Text Demo: Instead of visual demonstrations, this ablation provides textual input by listing observation vectors from a subsampled sequence of ten states from the demonstration, same as~\cite{peng2024adaptive}; 6) w/ Random Visual Demo: To test whether the VLM effectively extracts task-relevant information, we replace the high-quality visual demonstration input with random visual demonstrations (e.g., a falling Cartpole or Humanoid). The IRL phase, however, still utilizes the original high-quality demonstrations.
    
    % To further analyze the impact of the individual components of ELEMENTAL, we conduct ablation studies on two variants: 1) ELEMENTAL without Self-Reflection: This ablation removes the self-reflection mechanism introduced in Phase 3, keeping only the VLM-guided feature inference and policy learning phases; 2) ELEMENTAL without Visual Input: This ablation removes the visual input from the VLM, leaving only text-based language input for feature inference. The comparison with this ablation highlights the importance of visual demonstration in aligning robot behavior with the demonstration.
    % \vspace{-7pt}
\end{enumerate}\vspace{-7pt}
Performance is evaluated using an average task success rate of the final-100 steps during training, which is a more reliable metric versus only using the \emph{max} success during training reported by Eureka. In all experiments, we report the best performance across three random seeds to account for variable responses from GPT-4o. Hyperparameters and prompts are provided in supplementary. 

\subsection{Benchmarking Results}
\label{subsec:benchmarking_result}

\begin{table*}[t]
\centering
\footnotesize
\caption{This table shows reward correlation of inferred reward functions for EUREKA and ELEMENTAL (ours) on IsaacGym tasks. }
\label{tab:reward_correlation}
\begin{tabular}{lccccccccc}
\toprule
\multirow{2}{*}{Method} & \multicolumn{9}{c}{IsaacGym Environments} \\
\cmidrule(lr){2-10}
  & \multirow{ 2}{*}{Cartpole}  & Ball & \multirow{ 2}{*}{Quadcopter} & Franka & \multirow{ 2}{*}{Ant} & \multirow{ 2}{*}{Humanoid} & \multirow{ 2}{*}{Anymal} & Allegro& Shadow \\
    &  & Balance &  & Cabinet &  &  &  & Hand & Hand \\
\midrule
EUREKA & 0.77 & -0.53 & 0.96 & 0.93 & 1.00 & 0.59 & -0.86 & 0.34 & 0.00 \\
ELEMENTAL (Ours) & 0.99 & 0.85 & 0.89 & 0.98 & 1.00 & 0.98 & 0.97& 0.58 & 0.31 \\
\bottomrule
\end{tabular}
\end{table*}
% \vspace{-10pt}
\begin{table*}[t]
\footnotesize
\centering
\caption{This table compares the generalization performance of ELEMENTAL and EUREKA on Ant-variant environments. Results are maximized over three seeds. Bold denotes the best performance. }
\label{tab:generalization_max}
\begin{tabular}{lcccc}
\toprule
Method & Ant Original & w/ Reversed Obs & w/ Lighter Gravity & Ant Running Backward \\
\midrule
EUREKA & 6.88 & 5.96 & 4.39 & 5.62 \\
ELEMENTAL & \textbf{8.49} & \textbf{8.47} & \textbf{5.89} & \textbf{9.30} \\
w/o Visual Input & $8.16$ & $7.63$ & 3.14 & 7.46 \\
\bottomrule
\end{tabular}
\vspace{-10pt}
\end{table*}

In the first set of experiments, we benchmark ELEMENTAL and the baselines on the IsaacGym tasks, with the results presented in Table~\ref{tab:benchmark}. BC performs adequately on simpler tasks such as Cartpole and BallBalance, but fails on more complex tasks due to covariate shift. IRL without VLM-based feature extraction struggles to learn effectively in most tasks, highlighting the challenges posed by IsaacGym's high-dimensional state spaces. While EUREKA is able to learn capable policies, ELEMENTAL achieves on-average $42.3\%$ higher performance and outperforms EUREKA on eight out of nine tasks, demonstrating the effectiveness of integrating IRL with VLM-derived features and visual demonstration information. The learned reward correlation with ground-truth reward shown in Figure~\ref{tab:reward_correlation} also illustrates ELEMENTAL's strong ability to learn a well-aligned reward function by matching with demonstrations. 

\begin{figure}[t]
    \centering
    \includegraphics[width=\linewidth]{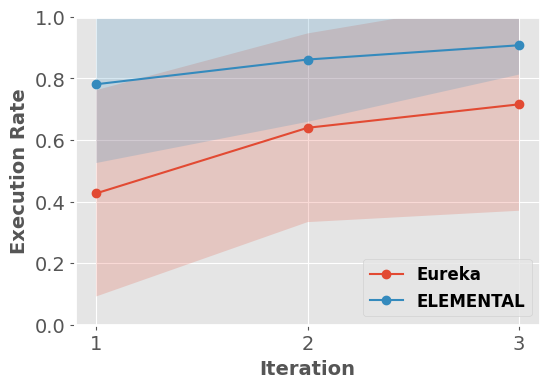}
    % \vspace{-20pt}
    \caption{A comparison of the code executation rate between ELEMENTAL and EUREKA in three iterations of the algorithms. }
    \vspace{5pt}
    \label{fig:execution_rate}
\end{figure}

To further evaluate whether VLM is more suitable for feature extraction or full reward function drafting, we examine the code execution rates of ELEMENTAL and EUREKA across three algorithm iterations. A higher code execution rate indicates fewer coding errors, suggesting better suitability for language models. As shown in Figure~\ref{fig:execution_rate}, ELEMENTAL achieves a successful code execution rate of approximately 80\% in the first iteration, compared to EUREKA's rate of less than 50\%. Although both algorithms improve with successive iterations, ELEMENTAL consistently generates more executable code. These results suggest that GPT-4o is more effective at feature extraction than drafting complete reward functions, supporting ELEMENTAL’s design choice to offload reward weighting to IRL.

The ablation results in Table~\ref{tab:benchmark} demonstrate the importance of ELEMENTAL’s key components. Removing self-reflection significantly reduces performance, confirming its role in refining both the feature function via VLM and the policy via IRL. Notably, even without self-reflection, ELEMENTAL outperforms standard IRL, indicating that the initial VLM-generated feature function already provides substantial benefits. The normalization steps are also crucial—removing them leads to unstable reward learning and inconsistent scaling, negatively impacting policy performance. The value of visual demonstrations is evident when comparing w/o Visual Input and w/ Text Demo: substituting visual demonstrations with subsampled state-vector text inputs, as in~\citet{peng2024adaptive}, degrades performance, supporting the hypothesis that VLMs better utilize visual inputs for feature extraction, particularly on tasks that are difficult to describe using natural language alone, such as FrankaCabinet and ShadowHand. Additionally, using random visual demonstrations leads to a significant drop in performance, confirming that VLMs extract meaningful task information when provided with high-quality demonstrations. These results highlight the necessity of self-reflection, normalization, and multimodal demonstrations in achieving ELEMENTAL’s stronger performance.

Under the same computational resource setup, EUREKA averaged 68.21 minutes of runtime across the nine tasks, while ELEMENTAL averaged 168.36 minutes. This increase is primarily due to the environment rollouts ELEMENTAL uses to estimate the reward gradient. However, this trade-off enables ELEMENTAL to learn a better-aligned reward function, leading to improved task success and generalization. We discuss future work to reduce runtime in Sec.~\ref{sec:limitation_future_work}.

\subsection{Generalization Results}
\label{subsec:generalization_result}

In the second set of experiments, we ask the question ``is LLM/VLM just remembering the reward function, as the knowledge cut-off date is Oct 2023, after IsaacGym is public?'' To answer this question, we test the generalization capability of ELEMENTAL by applying it to modified versions of the IsaacGym environments, particularly variants of the Ant task. For these modified environments, we change certain properties, such as state vector order, physics property (e.g., gravity coefficient), and the task, to evaluate whether ELEMENTAL’s VLM-driven feature inference combined with IRL can adapt to new, unseen environments better than EUREKA. We test on four Ant variants:
\begin{enumerate}[leftmargin=10pt]\vspace{-10pt}
\item Ant Original: The standard Ant task without modifications, serving as the baseline environment.\vspace{-8pt}
\item Ant with Reversed Observations: The order of the state vector is reversed, testing the algorithms' ability to adapt to changes in the structure of the input data.\vspace{-8pt}
\item Ant with Lighter Gravity: The gravity coefficient is reduced from $9.81$ to $5.00$ and requiring the features and policy to adjust for different dynamics. A performance drop is expected; the ant moves with less friction. \vspace{-8pt}
\item Ant Running Backward: The task is modified to require the Ant to move backward rather than forward, assessing how well the approaches generalize to a different task objective in the same environment.\vspace{-10pt}
\end{enumerate}
We show the comparison between ELEMENTAL and EUREKA in Table~\ref{tab:generalization_max}. In out-of-distribution tasks that language models have not seen in the training set, EUREKA's performance declines, suggesting GPT-4o might have memorized the IsaacGym task rewards, which is not helpful when the state vector is reversed, when the environment dynamics change, or when the task objective is altered. In contrast, ELEMENTAL queries the VLM only for feature functions—information not available in the IsaacGym public data—and uses the demonstration-matching IRL process to determine the reward weights. ELEMENTAL accomplishes an average performance improvement of 41.3\%, highlighting its robustness in adapting to changes in both the environment’s physical properties and the task's nature.

\section{Limitation and Future Work}
\label{sec:limitation_future_work}
While ELEMENTAL demonstrates strong improvements in LfD with VLMs, it has several limitations that open avenues for future research. One key limitation is its reliance on environment rollouts for IRL updates, which increases wall-clock running time as a trade-off that allows ELEMENTAL to infer reward functions that better align with demonstrations. Future work could explore more efficient IRL methods, such as Adversarial IRL (AIRL) or model-based RL, to reduce sample complexity. Additionally, parallelized or hardware-accelerated implementations could help mitigate runtime concerns, making ELEMENTAL more feasible for real-world deployment.

Another limitation is ELEMENTAL’s reliance on high-quality demonstrations. While our experiments show that ELEMENTAL benefits from high-quality visual demonstrations, its performance degrades when presented with suboptimal or random demonstrations. Future research could investigate methods that can filter or refine imperfect demonstrations. Additionally, incorporating human feedback in the self-reflection process could further refine feature selection and reward inference, making the framework more adaptable to diverse real-world conditions. Finally, given the demonstrated utility of visual inputs, alternative ways to provide demonstrations--such as continuous video demonstrations--could be explored to improve usability and generalization across different robotics tasks.

% \vspace{-5pt}
\section{Conclusion}
We introduced ELEMENTAL, a novel framework that integrates VLMs with LfD to draft feature functions from visual demonstrations. ELEMENTAL leverages IRL to match feature functions with demonstrations and introduces an iterative self-reflection mechanism for continuous improvement. Experiments showed that ELEMENTAL outperforms previous state-of-the-art methods by 42.3\% on standard IsaacGym tasks and 41.3\% in generalizing to novel tasks.

\section*{Impact Statement}
ELEMENTAL advances LfD by integrating VLMs with IRL to improve reward function engineering. Our contribution enables robots to learn from both natural language and visual demonstrations, reducing the need for hand-crafted reward functions and making robot learning more accessible to non-experts. By improving the alignment of learned behaviors with human demonstrations, ELEMENTAL has the potential to enhance human-robot interaction in areas such as assistive robotics, industrial automation, and adaptive learning systems.

Despite its benefits, ELEMENTAL, like all AI and robotics research, is a dual-use technology. One concern is the reliance on human demonstrations, which may introduce biases if training data is not diverse or representative. This could lead to unintended preferences or disparities in how robotic systems learn and generalize behaviors. Ensuring fair and inclusive demonstrations is essential for mitigating these risks. Additionally, while ELEMENTAL improves transparency in reward learning, its dependence on VLMs introduces potential challenges such as model hallucinations or overfitting to specific language priors.

Another ethical consideration is the potential misuse of ELEMENTAL for applications beyond its intended scope. While the framework is designed for positive use cases, such as making AI training more interpretable and user-aligned, the ability to iteratively refine reward functions could be leveraged in ways that raise ethical concerns, such as in surveillance or military applications. We advocate for responsible AI development and encourage the use of human-in-the-loop safeguards, bias auditing, and clear ethical guidelines for deployment.

Overall, ELEMENTAL represents a step toward more intuitive and generalizable AI systems by integrating demonstrations with language guidance. While its potential benefits are significant, careful attention must be given to ensuring fairness, transparency, and responsible use as these technologies continue to evolve.
\bibliography{example_paper}
\bibliographystyle{icml2025}

%%%%%%%%%%%%%%%%%%%%%%%%%%%%%%%%%%%%%%%%%%%%%%%%%%%%%%%%%%%%%%%%%%%%%%%%%%%%%%%
%%%%%%%%%%%%%%%%%%%%%%%%%%%%%%%%%%%%%%%%%%%%%%%%%%%%%%%%%%%%%%%%%%%%%%%%%%%%%%%
% APPENDIX
%%%%%%%%%%%%%%%%%%%%%%%%%%%%%%%%%%%%%%%%%%%%%%%%%%%%%%%%%%%%%%%%%%%%%%%%%%%%%%%
%%%%%%%%%%%%%%%%%%%%%%%%%%%%%%%%%%%%%%%%%%%%%%%%%%%%%%%%%%%%%%%%%%%%%%%%%%%%%%%
\newpage
\appendix
\onecolumn
\section{Prompts}
\begin{tcolorbox}[title=Initial System Prompt., colback=gray!5, colframe=black, colbacktitle=gray!20, coltitle=black, sharp corners, width=.99\linewidth,]
\begin{Verbatim}[breaklines=true]
You are a feature engineer trying to write relevant features for the reward function to solve learning-from-demonstration (inverse reinforcement learning) tasks as effective as possible.

Your goal is to write a feature function for the environment that will help the agent construct a linear reward function with the constructed features via inverse reinforcement learning to accomplish the task described in text and the demonstration.

Your feature function should use useful variables from the environment as inputs. The feature function signature must follow:
@torch.jit.script
def compute_feature(obs_buf: torch.Tensor) -> Dict[str, torch.Tensor]:
    ...
    return {}

Since the feature function will be decorated with @torch.jit.script, please make sure that the code is compatible with TorchScript (e.g., use torch tensor instead of numpy array).

You should not wrap the function within a class.

Make sure any new tensor or variable you introduce is on the same device as the input tensors. 
\end{Verbatim}
\end{tcolorbox}

\begin{tcolorbox}[title=Initial User Prompt., colback=gray!5, colframe=black, colbacktitle=gray!20, coltitle=black, sharp corners, width=.99\linewidth,]
\begin{Verbatim}[breaklines=true]
The Python environment is
{task_obs_code_string}

Write a feature function for the following task: {task_description}.

Three keyframes of a demonstration for how to accomplish the task are shown in the image (superimposing agent pos).
\end{Verbatim}
\end{tcolorbox}

\begin{tcolorbox}[title=Code output instruction., colback=gray!5, colframe=black, colbacktitle=gray!20, coltitle=black, sharp corners, width=.99\linewidth,]
\begin{Verbatim}[breaklines=true]
The input of the feature function is a torch.Tensor named `obs_buf` that is a batched state (shape: [batch, num_obs]).

The output of the feature function should be a dictionary where the keys are the names of the features and the values are the corresponding feature values for the input state.

You must respect the function signature.

The code output should be formatted as a python code string: "```python ... ```".

Some helpful tips for writing the feature function code:

    (1) You may find it helpful to normalize the features to a fixed range by applying transformations
    
    (2) The feature code's input variables must be obs_buf: torch.Tensor, which corresponds to the state observation (self.obs_buf) returned by the environment compute_observations() function. Under no circumstance can you introduce new input variables.
    
    (3) Each output feature should only one a single dimension (shape: [batch]).
    
    (4) You should think step-by-step: first, think what is important in the task based on the task description and the demonstration and come up with names of the features, then, write code to calculate each feature
    
    (5) You should be aware that the downstream inverse reinforcement learning only creates reward functions that are linear function of the constructed features; thus, it is important to construct expressive features that humans do care in this task
    
    (6) Do not use unicode anywhere such as \u03c0 (pi)
\end{Verbatim}
\end{tcolorbox}

\begin{tcolorbox}[title=Self-Reflection prompt., colback=gray!5, colframe=black, colbacktitle=gray!20, coltitle=black, sharp corners, width=.99\linewidth,]
\begin{Verbatim}[breaklines=true]
We trained reward and policy via inverse reinforcement learning using the provided feature function code with the demonstration.

We tracked the feature values as well as episode lengths.

The mean values of the last {eval_avg_horizon} steps from the learned policy are:
{insert}

Please carefully analyze the feedback and provide a new, improved feature function that can better solve the task. Some helpful tips for analyzing the feedback:

    (1) If the episode lengths are low, it likely means the policy is unsuccessful
    
    (2) If the feature counts are significantly different between demo and learned behavior, then this means IRL cannot match this feature with the demo as it is written. You may consider
    
        (a) Change its scale
        
        (b) Re-writing the feature: check error in the feature computation (e.g., indexing the observation vector) and be careful about outlier values that may occur in the computation
        
        (c) Discarding the feature
        
    (3) If a feature has near-zero weight, the feature may be unimportant. You can consider discarding the feature or rewriting it.
    
    (4) You may add/remove features as you see appropriate.
    
Please analyze each existing features in the suggested manner above first, and then write the feature function code.
\end{Verbatim}
\end{tcolorbox}

\section{ELEMENTAL Hyerparameters}
We tune hyperparameters via a grid search. We summarize the ELEMENTAL hyperparameters in Table~\ref{tab:hyperparameters}. All other hyperparameters follow EUREKA's default setup. 

\begin{table}[h]
\centering
\begin{tabular}{cc}
\hline
\textbf{Hyperparameters} & \textbf{Value} \\
\hline
Reward learning rate $\alpha$ & $1.0$ \\
Approximate MaxEnt-IRL number of iterations $m$ & 5 \\
Policy training steps $k$ & 500 \\
Number of algorithm iterations & 3 \\
Code samples generated per iteration & 1 for ShadowHand, 3 for all other tasks \\
Policy Neural Network Architecture & Fully-connected [32, 32] with ReLU activation \\
\hline
\end{tabular}
\caption{Hyperparameters and their values}
\label{tab:hyperparameters}
\end{table}
% \bibliography{iclr2025_conference}
% \bibliographystyle{iclr2025_conference}

\section{Detailed Restuls}
\begin{table}[t]
\footnotesize
\centering
\caption{This table compares generalization performance of ELEMENTAL and EUREKA on Ant-variant environments. Results (mean$\pm$std) are averaged over three seeds. Bold denotes the best performance. }
\label{tab:generalization}
\begin{tabular}{lcccc}
\toprule
Method & Ant Original & w/ Reversed Obs & w/ Lighter Gravity & Ant Running Backward \\
\midrule
EUREKA & 4.44$\pm 2.45$ & $4.11\pm 2.07$ & 2.94$\pm 2.53$ & 3.51$\pm 3.03$ \\
ELEMENTAL & \textcolor{black}{6.80$\pm 1.17$} & \textcolor{black}{7.40$\pm 0.98$} & 4.35\textcolor{black}{$\pm 1.20$} & 7.41\textcolor{black}{$\pm 1.25$} \\
w/o Visual Input & $7.03\pm 1.67$ & $7.07\pm1.07$ & $2.57\pm0.68$ & $6.02\pm2.03$ \\
\bottomrule
\end{tabular}
\end{table}
\subsection{Generalization experiment}
The mean and standard deviation for generalization experiment across three seeds are shown in Table~\ref{tab:generalization}. 

\subsection{Case Study}
In this subsection, we present a case study illustrating the iterative process of ELEMENTAL on the Humanoid task. The initial feature function drafted by the Vision-Language Model (VLM) is shown in Box~\ref{box:1}. The proposed features—forward\_velocity, uprightness, and heading\_alignment—are well-aligned with the task objectives of running efficiently while maintaining balance and direction. These features provide a strong starting point for the learning process. 

Using this initial feature function, ELEMENTAL trains the IRL process, calculates the feature counts for both the generated trajectories and the demonstration, and feeds this feedback back to the VLM, as shown in Box~\ref{box:2}. The feedback reveals key discrepancies, such as lower forward\_velocity. Based on this analysis, the VLM revises the feature function, as shown in Box~\ref{box:3}. Notably, the revised function introduces a new feature, lateral\_velocity, which captures stride consistency by taking the absolute value of the lateral movement. This demonstrates the VLM’s capability to construct nonlinear features, expanding the expressiveness of the feature function.

Finally, ELEMENTAL trains the IRL process again using the updated feature function and compiles the feedback. The resulting reward weights assign the highest importance to forward\_velocity, with relatively minor contributions from stability-related objectives such as uprightness and lateral\_velocity. This distribution aligns well with human intuition for the task, where speed is the primary objective, and stability features serve as secondary constraints.

This case study highlights the potential of ELEMENTAL to produce models at every stage of its pipeline. The feature functions generated by the VLM are human-readable and meaningful, allowing practitioners to inspect and refine them as needed. Additionally, the linear weights learned by ELEMENTAL during the IRL process indicate the relative importance of each feature in matching the demonstration. The iterative self-reflection mechanism ensures that the learned policies and reward functions evolve in alignment with the demonstration.

\begin{prompt}[title=Box \thetcbcounter: 1st round VLM result, colback=gray!5, colframe=black, colbacktitle=gray!20, coltitle=black, sharp corners, width=.99\linewidth,float,floatplacement=t,label=box:1]
\begin{tiny}
\begin{Verbatim}[breaklines=true]
To create a feature function aimed at helping a humanoid robot run as fast as possible, we should focus on features that are most relevant to running speed and ensuring the humanoid's balance and stability. Useful features could include the forward velocity, balance (uprightness), and perhaps a component that ensures the humanoid maintains forward momentum without falling.

Here's the feature function based on the observation buffer structure:

```python
import torch
from typing import Dict

@torch.jit.script
def compute_feature(obs_buf: torch.Tensor) -> Dict[str, torch.Tensor]:
    # Extract relevant features from the observation buffer
    # Assuming obs_buf shape: [batch_size, num_obs]
    
    # Feature: forward velocity (x direction in local frame)
    forward_velocity = obs_buf[:, 1]
    
    # Feature: uprightness (use the up projection)
    uprightness = obs_buf[:, 10]
    
    # Feature: heading alignment (use the heading projection)
    heading_alignment = obs_buf[:, 11]
    
    # Normalize the features
    # Assuming value normalizations such as the velocity range, uprightness [-1, 1], heading alignment [-1, 1]
    max_forward_velocity = 10.0  # This is an arbitrary normalization
    forward_velocity = forward_velocity / max_forward_velocity  # Normalizing forward velocity

    # Combine features into output dictionary
    features = {
        "forward_velocity": forward_velocity,
        "uprightness": uprightness,
        "heading_alignment": heading_alignment,
    }
    
    return features
```

### Explanation:

1. **`forward_velocity`**: The observation buffer's second column (`obs_buf[:, 1]`) captures the forward velocity in the local frame.

2. **`uprightness`**: The observation buffer's 11th column (`obs_buf[:, 10]`) captures the upright projection, which is crucial to ensure the humanoid isn't falling over.

3. **`heading_alignment`**: The observation buffer's 12th column (`obs_buf[:, 11]`) captures heading alignment, which helps ensure the humanoid is moving forward in the correct direction.

These features will enable the inverse reinforcement learning process to construct a reward function that emphasizes running fast while maintaining balance and a straight heading.

\end{Verbatim}
\end{tiny}
\end{prompt}

\begin{prompt}[title=Box \thetcbcounter: 1st round feedback, colback=gray!5, colframe=black, colbacktitle=gray!20, coltitle=black, sharp corners, width=.99\linewidth,float,floatplacement=t,label=box:2]
\begin{tiny}
\begin{Verbatim}[breaklines=true]
Calculating mean features on the demonstration results in the following: 
episode_lengths: 1000.0 
forward_velocity: 0.57 
uprightness: 0.94 
heading_alignment: 0.91 
We trained reward and policy via inverse reinforcement learning using the provided feature function code with the demonstration.
We tracked the feature values as well as episode lengths.
The mean values of the last 100 steps from the learned policy are:
forward_velocity: 0.4 
uprightness: 0.99 
heading_alignment: 0.99 
IRL reward: 0.05 
episode_lengths: 691.47 
IRL feature weights: {'forward_velocity': 0.693, 'uprightness': -0.107, 'heading_alignment': -0.201}
Please carefully analyze the feedback and provide a new, improved feature function that can better solve the task. Some helpful tips for analyzing the feedback:
    (1) If the episode lengths are low, it likely means the policy is unsuccessful
    (2) If the feature counts are significantly different between demo and learned behavior, then this means IRL cannot match this feature with the demo as it is written. You may consider
        (a) Change its scale
        (b) Re-writing the feature: check error in the feature computation (e.g., indexing the observation vector) and be careful about outlier values that may occur in the computation
        (c) Discarding the feature
    (3) If a feature has near-zero weight, the feature may be unimportant. You can consider discarding the feature or rewriting it.
    (4) You may add/remove features as you see appropriate.
Please analyze each existing features in the suggested manner above first, and then write the feature function code.The input of the feature function is a torch.Tensor named `obs_buf` that is a batched state (shape: [batch, num_obs]).
The output of the feature function should be a dictionary where the keys are the names of the features and the values are the corresponding feature values for the input state.
You must respect the function signature.
The code output should be formatted as a python code string: "```python ... ```".

Some helpful tips for writing the feature function code:
    (1) You may find it helpful to normalize the features to a fixed range by applying transformations
    (2) The feature code's input variables must be obs_buf: torch.Tensor, which corresponds to the state observation (self.obs_buf) returned by the environment compute_observations() function. Under no circumstance can you introduce new input variables.
    (3) Each output feature should only one a single dimension (shape: [batch]).
    (4) You should think step-by-step: first, think what is important in the task based on the task description and the demonstration and come up with names of the features, then, write code to calculate each feature
    (5) You should be aware that the downstream inverse reinforcement learning only creates reward functions that are linear function of the constructed features; thus, it is important to construct expressive features that humans do care in this task
    (6) Do not use unicode anywhere such as \u03c0 (pi)

\end{Verbatim}
\end{tiny}
\end{prompt}

\begin{prompt}[title=Box \thetcbcounter: 2nd round VLM result, colback=gray!5, colframe=black, colbacktitle=gray!20, coltitle=black, sharp corners, width=.99\linewidth,float,floatplacement=t,label=box:3]
\begin{tiny}
\begin{Verbatim}[breaklines=true]
Based on the provided feedback and analysis of how the current features are performing, we can observe a few key points:

1. **Forward Velocity:**
   - The policy is achieving a lower forward velocity than in the demonstration. This suggests that the current method of calculating or normalizing this feature might need adjustment.
   - The forward velocity weight is quite significant (`0.693`). 

2. **Uprightness:**
   - The policy has a higher uprightness than the demonstration. The weight for uprightness is negative (`-0.107`), which may indicate that the model learns to be overly cautious about staying upright, potentially at the expense of moving fast. This negative weight suggests we might need to rethink how this feature is influencing the reward.
  
3. **Heading Alignment:**
   - The policy has a higher heading alignment than the demonstration. Similar to uprightness, the weight is also negative and relatively significant (`-0.201`), indicating it might be overemphasizing this feature.

Given these points, we aim to improve the feature function by:
1. Adjusting the scaling or normalization of the forward velocity.
2. Revisiting and potentially revising the uprightness and heading alignment features to better match the demonstration characteristics.
3. Considering new or additional features that might encourage the desired behavior such as stability or stride consistency.

Here is the improved feature function:

```python
import torch
from typing import Dict

@torch.jit.script
def compute_feature(obs_buf: torch.Tensor) -> Dict[str, torch.Tensor]:
    # Extract relevant features from the observation buffer
    # Assuming obs_buf shape: [batch_size, num_obs]
    
    # Feature: forward velocity (x direction in local frame)
    forward_velocity = obs_buf[:, 1]
    
    # Adjust scaling of forward velocity to emphasize higher speeds, assuming range [-1, 1]
    max_forward_velocity = 1.0  # New scaling factor to more accurately reflect achievable max speeds
    forward_velocity = forward_velocity / max_forward_velocity  # Normalizing forward velocity

    # Feature: uprightness (use the up projection)
    uprightness = obs_buf[:, 10]
    
    # Re-normalize uprightness to give more incentive to be upright but not be overly conservative
    uprightness = (uprightness - 0.5) * 2  # Adjusting the range to new scaling [-1, 1]

    # Feature: heading alignment (use the heading projection)
    heading_alignment = obs_buf[:, 11]
    
    # Normalize heading alignment to confirm alignment direction
    heading_alignment = (heading_alignment - 0.5) * 2  # Adjusting the range similarly

    # Introducing a new feature: lateral velocity (to ensure stability)
    lateral_velocity = obs_buf[:, 2]
    lateral_velocity = torch.abs(lateral_velocity)  # Use absolute value to avoid penalizing/stabilize sideway motions

    # Normalize lateral velocity, supposing it ranges [-1, 1]
    max_lateral_velocity = 1.0
    lateral_velocity = lateral_velocity / max_lateral_velocity
    
    # Combine features into output dictionary
    features = {
        "forward_velocity": forward_velocity,
        "uprightness": uprightness,
        "heading_alignment": heading_alignment,
        "lateral_velocity": lateral_velocity
    }
    
    return features
```

### Explanation:

1. **`forward_velocity`**: Adjusted scaling to make achievable velocities more impactful.
2. **`uprightness`**: Adjusted to a new normalization to try to balance upright positioning without making the humanoid overly cautious.
3. **`heading_alignment`**: Similar adjustment for normalization.
4. **`lateral_velocity`**: Added a new feature to penalize sideways movement, ensuring stability and straightforward motion.

These changes aim to provide a more nuanced reward function, hopefully promoting behaviors closer to the demonstration while balancing speed and stability.
\end{Verbatim}
\end{tiny}
\end{prompt}

\begin{prompt}[title=Box \thetcbcounter: 2nd round feedback, colback=gray!5, colframe=black, colbacktitle=gray!20, coltitle=black, sharp corners, width=.99\linewidth,float,floatplacement=t,label=box:4]
\begin{tiny}
\begin{Verbatim}[breaklines=true]
Calculating mean features on the demonstration results in the following: 
episode_lengths: 1000.0 
forward_velocity: 5.67 
uprightness: 0.89 
heading_alignment: 0.83 
lateral_velocity: 2.31 
We trained reward and policy via inverse reinforcement learning using the provided feature function code with the demonstration.
We tracked the feature values as well as episode lengths.
The mean values of the last 100 steps from the learned policy are:
forward_velocity: 5.04 
uprightness: 0.65 
heading_alignment: 0.92 
lateral_velocity: 2.05 
IRL reward: 3.54 
episode_lengths: 932.42 
IRL feature weights: {'forward_velocity': 0.549, 'uprightness': 0.253, 'heading_alignment': -0.079, 'lateral_velocity': 0.119}
Please carefully analyze the feedback and provide a new, improved feature function that can better solve the task. Some helpful tips for analyzing the feedback:
    (1) If the episode lengths are low, it likely means the policy is unsuccessful
    (2) If the feature counts are significantly different between demo and learned behavior, then this means IRL cannot match this feature with the demo as it is written. You may consider
        (a) Change its scale
        (b) Re-writing the feature: check error in the feature computation (e.g., indexing the observation vector) and be careful about outlier values that may occur in the computation
        (c) Discarding the feature
    (3) If a feature has near-zero weight, the feature may be unimportant. You can consider discarding the feature or rewriting it.
    (4) You may add/remove features as you see appropriate.
Please analyze each existing features in the suggested manner above first, and then write the feature function code.The input of the feature function is a torch.Tensor named `obs_buf` that is a batched state (shape: [batch, num_obs]).
The output of the feature function should be a dictionary where the keys are the names of the features and the values are the corresponding feature values for the input state.
You must respect the function signature.
The code output should be formatted as a python code string: "```python ... ```".

Some helpful tips for writing the feature function code:
    (1) You may find it helpful to normalize the features to a fixed range by applying transformations
    (2) The feature code's input variables must be obs_buf: torch.Tensor, which corresponds to the state observation (self.obs_buf) returned by the environment compute_observations() function. Under no circumstance can you introduce new input variables.
    (3) Each output feature should only one a single dimension (shape: [batch]).
    (4) You should think step-by-step: first, think what is important in the task based on the task description and the demonstration and come up with names of the features, then, write code to calculate each feature
    (5) You should be aware that the downstream inverse reinforcement learning only creates reward functions that are linear function of the constructed features; thus, it is important to construct expressive features that humans do care in this task
    (6) Do not use unicode anywhere such as \u03c0 (pi)

\end{Verbatim}
\end{tiny}
\end{prompt}
%%%%%%%%%%%%%%%%%%%%%%%%%%%%%%%%%%%%%%%%%%%%%%%%%%%%%%%%%%%%%%%%%%%%%%%%%%%%%%%
%%%%%%%%%%%%%%%%%%%%%%%%%%%%%%%%%%%%%%%%%%%%%%%%%%%%%%%%%%%%%%%%%%%%%%%%%%%%%%%

\end{document}